\documentclass[11pt,letterpaper]{article}
\usepackage{emnlp2017}
\usepackage{times}
\usepackage{latexsym}
\usepackage{listings}
\usepackage{footmisc}
\usepackage{textcomp}
\usepackage{graphicx} 
\usepackage{enumitem}
\usepackage{booktabs} 

\emnlpfinalcopy

\def\emnlppaperid{469}

\makeatletter
\def\@xfootnote[#1]{%
\protected@xdef\@thefnmark{#1}%
\@footnotemark\@footnotetext}
\makeatother

\title{Sharp Models on Dull Hardware: Fast and Accurate Neural Machine Translation Decoding on the CPU}

\author{Jacob Devlin \\
{\tt jdevlin@microsoft.com} \\ Microsoft Research}

\date{}

\begin{document}

\maketitle

\begin{abstract}
Attentional sequence-to-sequence models have become the new standard for machine translation, but one challenge of such models is a significant increase in training and decoding cost compared to phrase-based systems. Here, we focus on efficient decoding, with a goal of achieving accuracy close the state-of-the-art in neural machine translation (NMT), while achieving CPU decoding speed/throughput close to that of a phrasal decoder.

We approach this problem from two angles: First, we describe several techniques for speeding up an NMT beam search decoder, which obtain a 4.4x speedup over a very efficient baseline decoder without changing the decoder output. Second, we propose a simple but powerful network architecture which uses an RNN (GRU/LSTM) layer at bottom, followed by a series of stacked fully-connected layers applied at every timestep. This architecture achieves similar accuracy to a deep recurrent model, at a small fraction of the training and decoding cost. By combining these techniques, our best system achieves a very competitive accuracy of 38.3 BLEU on WMT English-French NewsTest2014, while decoding at 100 words/sec on single-threaded CPU. We believe this is the best published accuracy/speed trade-off of an NMT system.
\end{abstract}

\section{Introduction}

Attentional sequence-to-sequence models have become the new standard for machine translation over the last two years, and with the unprecedented improvements in translation accuracy comes a new set of technical challenges. One of the biggest challenges is the high training and decoding costs of these neural machine translation (NMT) system, which is often at least an order of magnitude higher than a phrase-based system trained on the same data. For instance, phrasal MT systems were able achieve single-threaded decoding speeds of 100-500 words/sec on decade-old CPUs \cite{quirk2007}, while \citet{jean2015} reported single-threaded decoding speeds of 8-10 words/sec on a shallow NMT system. \citet{googlenmt} was able to reach CPU decoding speeds of 100 words/sec for a deep model, but used 44 CPU cores to do so. There has been recent work in speeding up decoding by reducing the search space \cite{kim2016}, but little in computational improvements.

In this work, we consider a production scenario which requires low-latency, high-throughput NMT decoding. We focus on CPU-based decoders, since GPU/FPGA/ASIC-based decoders require specialized hardware deployment and logistical constraints such as batch processing. Efficient CPU decoders can also be used for on-device mobile translation. We focus on single-threaded decoding and single-sentence processing, since multiple threads can be used to reduce latency but not total throughput.

We approach this problem from two angles: In Section~\ref{sec:speedups}, we describe a number of techniques for improving the speed of the decoder, and obtain a 4.4x speedup over a highly efficient baseline. These speedups do not affect decoding results, so they can be applied universally. In Section~\ref{sec:model_improvements}, we describe a simple but powerful network architecture which uses a single RNN (GRU/LSTM) layer at the bottom with a large number of fully-connected (FC) layers on top, and obtains improvements similar to a deep RNN model at a fraction of the training and decoding cost.

\section{Data Set}
The data set we evaluate on in this work is WMT English-French NewsTest2014, which has 380M words of parallel training data and a 3003 sentence test set. The NewsTest2013 set is used for validation. In order to compare our architecture to past work, we train a word-based system without any data augmentation techniques. The network architecture is very similar to \citet{bahdanau2014}, and specific details of layer size/depth are provided in subsequent sections. We use an 80k source/target vocab and perform standard unk-replacement \cite{jean2015} on out-of-vocabulary words. Training is performed using an in-house toolkit.

\section{Baseline Decoder}
Our baseline decoder is a standard beam search decoder \cite{sutskever2014} with several straightforward performance optimizations:
\begin{itemize}[noitemsep]
\item It is written in pure C++, with no heap allocation done during the core search.
\item A candidate list is used to reduce the output softmax from 80k to \texttildelow$500$. We run word alignment \cite{brown1993} on the training and keep the top 20 context-free translations for each source word in the test sentence.
\item The Intel MKL library is used for matrix multiplication, as it is the fastest floating point matrix multiplication library for CPUs.
\item Early stopping is performed when the top partial hypothesis has a log-score of $\delta = 3.0$ worse than the best completed hypothesis.
\item Batching of matrix multiplication is applied when possible. Since each sentence is decoded separately, we can only batch over the hypotheses in the beam as well as the input vectors on the source side.
\end{itemize}

\section{Decoder Speed Improvements}
\label{sec:speedups}

This section describes a number of speedups that can be made to a CPU-based attentional sequence-to-sequence beam decoder. Crucially, none of these speedups affect the actual mathematical computation of the decoder, so they can be applied to any network architecture with a guarantee that they will not affect the results.\footnote{Some speedups apply quantization which leads to small random perturbations, but these change the BLEU score by less than 0.02.}

The model used here is similar to the original implementation of \citet{bahdanau2014}. The exact target GRU equation is:
{\vskip -0.18in}
{\fontsize{10.0}{10.0}\selectfont
\begin{eqnarray*}
d_{ij} & = & {\rm tanh}(W_a{h_{i-1}} + V_a{x_i}){\cdot}{\rm tanh}(U_as_j) \\
\alpha_{ij} & = & \frac{e^{d_{ij}}}{\sum_{j'}e^{d_{ij'}}} \\
c_{i} &=& \sum_{j} \alpha_{ij}s_j \\
u_i & = & \sigma(W_u{h_{i-1}} + V_u{x_i} + U_u{c_i} + b_u) \\
r_i & = & \sigma(W_r{h_{i-1}} + V_r{x_i} + U_r{c_i} + b_r) \\
\hat{h}_i & = & \sigma(r_i{\odot}(W_h{h_{i-1}}) + V_h{x_i} + U_h{c_i} + b_h) \\
h_i & = & u_ih_{i-1} + (1 - u_i)\hat{h}_i
\end{eqnarray*}
}
Where $W_*$, $V_*$, $U_*$, $b_*$ are learned parameters, $s_j$ is the hidden vector of the $j^{\rm th}$ source word, $h_{i-1}$ is the previous target recurrent vector, $x_i$ is the target input (e.g., embedding of previous word).

We also denote the various hyperparameters: $b$ for the beam size, $r$ for the recurrent hidden size, $e$ is the embedding size, $|S|$ for the source sentence length, and $|T|$ for the target sentence length, $|E|$ is the vocab size.

\subsection{16-Bit Matrix Multiplication}
Although CPU-based matrix multiplication libraries are highly optimized, they typically only operate on 32/64-bit floats, even though DNNs can almost always operate on much lower precision without degredation of accuracy \cite{han2015}. However, low-precision math (1-bit to 7-bit) is difficult to implement efficiently on the CPU, and even 8-bit math has limited support in terms of vectorized (SIMD) instruction sets. Here, we use 16-bit fixed-point integer math, since it has first-class SIMD support and requires minimal changes to training. Training is still performed with 32-bit floats, but we clip the weights to the range [-1.0, 1.0] the {\tt relu} activation to [0.0, 10.0] to ensure that all values fit into 16-bits with high precision. A reference implementation of 16-bit multiplication in C++/SSE2 is provided in the supplementary material, with a thorough description of low-level details.\footnote{Included as ancillary file in Arxiv submission, on right side of submission page.}

\begin{figure}[thb]
\begin{center}
\includegraphics[width=225px]{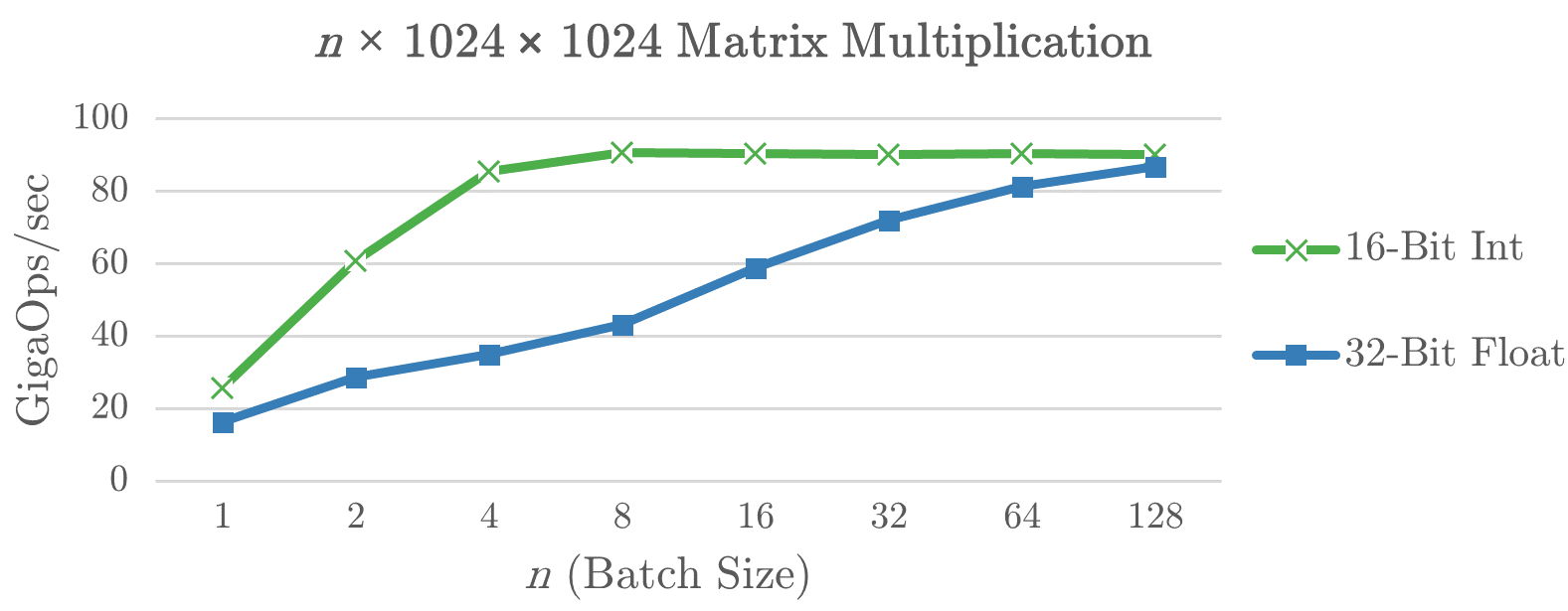}
{\vskip -0.1in}
\caption{{\small Single-threaded matrix multiplication using our 16-bit fixed-point vs. Intel MKL's 32-bit float, averaged over 10,000 multiplications. Both use the AVX2 instruction set.
}}
{\vskip -0.1in}
\label{fig:matrix_mult}
\end{center}
\end{figure}

A comparison between our 16-bit integer implementation and Intel MKL's 32-bit floating point multiplication is given in Figure~\ref{fig:matrix_mult}. We can see that 16-bit multiplication is 2x-3x faster than 32-bit multiplication for batch sizes between 2 and 8, which is the typical range of the beam size $b$. We are able to achieve greater than a 2x speedup in certain cases because we pre-process the weight matrix offline to have optimal memory layout, which is a capability BLAS libraries do not have.

\label{sec:matrix_mult}

\subsection{Pre-Compute Embeddings}
In the first hidden layer on the source and target sides, $x_i$ corresponds to word embeddings. Since this is a closed set of values that are fixed after training, the vectors $V{x_i}$ can be pre-computed \cite{devlin2014} for each word in the vocabulary and stored in a lookup table. This can only be applied to the first hidden layer.

Pre-computation does increase the memory cost of the model, since we must store $r \times 3$ floats per word instead of $e$. However, if we only compute the $k$ most frequently words (e.g., $k = 8,000$), this reduces the pre-computation memory by 90\% but still results in 95\%+ token coverage due to the Zipfian distribution of language.

\subsection{Pre-Compute Attention}
The attention context computation in the GRU can be re-factored as follows: 
{
\setlength{\belowdisplayskip}{2pt} \setlength{\belowdisplayshortskip}{2pt}
\setlength{\abovedisplayskip}{2pt} \setlength{\abovedisplayshortskip}{2pt}

\[U{c_i} = U(\sum_j \alpha_{ij}s_j) = \sum_j \alpha_{ij}(Us_j)\]
}
Crucially, the hidden vector representation $s_j$ is only dependent on the source sentence, while $a_{ij}$ is dependent on the target hypothesis. Therefore, the original computation $U{c_i}$ requires total $|T| \times b$ multiplications per sentence, but the re-factored version $Us_j$ only requires total $|S|$ multiplications. The expectation over $\alpha$ must still be computed at each target timestep, but this is much less expensive than the multiplication by $U$. 

\subsection{SSE \& Lookup Tables}
For the element-wise vector functions use in the GRU, we can use vectorized instructions (SSE/AVX) for the {\tt add} and {\tt multiply} functions, and lookup tables for {\tt sigmoid} and {\tt tanh}. Reference implementations in C++ are provided in the supplementary material.

\subsection{Merge Recurrent States}
In the GRU equation, for the first target hidden layer, $x_i$ represents the previously generated word, and $h_{i-1}$ encodes the hypothesis up to {\it two} words before the current word. Therefore, if two partial hypotheses in the beam only differ by the last emitted word, their $h_{i-1}$ vectors will be identical. Thus, we can perform matrix multiplication $Wh_{i-1}$ only on the {\it unique} $h_{i-1}$ vectors in the beam at each target timestep. For a beam size of $b = 6$, we measured that the ratio of unique $h_{i-1}$ compared to total $h_{i-1}$ is approximately 70\%, averaged over several language pairs. This can only be applied to the first target hidden layer.

\begin{table}[thb]
\begin{center}
\begin{tabular}{lcc}
\toprule
& \textbf{Words/Sec.} & \textbf{Speedup} \\
\textbf{Type}    & {\fontsize{8.0}{8.0}\selectfont  \textbf{(Single-Threaded)}} & \textbf{Factor} \\
\midrule
Baseline & 95    & 1.00x \\
+ 16-Bit Mult.   & 248 & 2.59x \\
+ Pre-Comp. Emb. & 311 & 3.25x \\
+ Pre-Comp. Att. & 342 & 3.57x \\
+ SSE \& Lookup  & 386 & 4.06x \\
+ Merge Rec.     & 418 & 4.37x \\
\bottomrule
\end{tabular}
\label{table:seq_to_seq_results}
\end{center}
{\vskip -0.1in}
\caption{{\small Decoding speeds on an Intel E5-2660 CPU, processing each sentence independently.}}
\label{table:speedup_results}
{\vskip -0.1in}
\end{table}

\subsection{Speedup Results}
Cumulative results from each of the preceding speedups are presented in Table~\ref{table:speedup_results}, measured on WMT English-French NewsTest2014. The NMT architecture evaluated here uses 3-layer 512-dimensional bidirectional GRU for the source, and a 1-layer 1024-dimensional attentional GRU for the target. Each sentence is decoded independently with a beam of 6. Since these speedups are all mathematical identities excluding quantization noise, all outputs achieve 36.2 BLEU and are 99.9\%+ identical.

The largest improvement is from 16-bit matrix multiplication, but all speedups contribute a significant amount. Overall, we are able to achieve a 4.4x speedup over a fast baseline decoder. Although the absolute speed is impressive, the model only uses one target layer and is several BLEU behind the SOTA, so the next goal is to maximize model accuracy while still achieving speeds greater than some target, such as 100 words/sec.

\begin{table*}[thb]
{\fontsize{10.5}{10.1}\selectfont
\begin{center}
\begin{tabular}{lcc}
\toprule
& & \\ \noalign{\vskip -0.1in}
& & \textbf{Words/Sec} \\
\textbf{System} & \textbf{BLEU} & {\fontsize{8.0}{8.0}\selectfont \textbf{(Single-Threaded)}} \\
\midrule 
& & \\ \noalign{\vskip -0.1in}
Basic Phrase-Based MT \citep{schwenk2014} & 33.1 & - \\
SOTA Phrase-Based MT \citep{durrani2014} & 37.0 & - \\
6-Layer Non-Attentional Seq-to-Seq LSTM \citep{luong2014} & 33.1 & - \\
RNN Search, 1-Layer Att. GRU, w/ Large Vocab \citep{jean2015} & 34.6 & $\dagger$ \\
Google NMT, 8-Layer Att. LSTM, Word-Based \citep{googlenmt} & 37.9 & $\flat$ \\
Google NMT, 8-Layer Att. LSTM, WPM-32k \citep{googlenmt} & 39.0$^\ddagger$ & $\flat$ \\
Baidu Deep Attention, 8-Layer Att. LSTM \cite{zhou2016} & 39.2 & - \\
\midrule
& & \\ \noalign{\vskip -0.1in}
(S1) Trg: 1024-AttGRU & 36.2 & 418 \\
(S2) Trg: 1024-AttGRU + 1024-GRU & 36.8 & 242 \\
(S3) Trg: 1024-AttGRU + 3-Layer 768-FC-Relu + 1024-FC-Tanh & 37.1 & 271 \\
(S4) Trg: 1024-AttGRU + 7-Layer 768-FC-Relu + 1024-FC-Tanh & 37.4 & 229 \\
(S5) Trg: 1024-AttGRU + 7-Layer 768-FC-Relu + 1024-GRU & 37.6 & 157 \\
(S6) Trg: 1024-AttGRU + 15-Layer 768-FC-Relu + 1024-FC-Tanh & 37.3 & 163 \\
(S7) Src: 8-Layer LSTM, Trg: 1024-AttLSTM + 7-Layer 1024-LSTM$^\S$ & 37.8 & 28 \\
\midrule
& & \\ \noalign{\vskip -0.1in}
\textbf{(E1) Ensemble of 2x Model (S4)} & \textbf{38.3} & \textbf{102} \\
(E2) Ensemble of 3x Model (S4) & 38.5 & 65 \\
\bottomrule
\end{tabular}
\label{table:seq_to_seq_results}
\end{center}
\caption{{\small Results on WMT English-French NewsTest2014. Models (S1)-(S6) use a 3-layer 512-dim bidirectional GRU for the source side. The CPU is an Intel Haswell E5-2660. $\dagger$ Reported as {\texttildelow}8 words/sec on one CPU core. $\flat$ Reported as {\texttildelow}100 words/sec, parallelized across 44 CPU cores. $\ddagger$ Uses word-piece tokenization, all others are word-based. $\S$ Reproduction of Google NMT, Word-Based.}}
\label{table:model_results}
}
\end{table*}

\section{Model Improvements}

\label{sec:model_improvements}
In NMT, like in many other deep learning tasks, accuracy can be greatly improved by adding more hidden layers, but training and decoding time increase significantly \cite{luong2014, zhou2016, googlenmt}. Several past works have noted that convolutional neural networks (CNNs) are significantly less expensive than RNNs, and replaced the source and/or target side with a CNN-based architecture \cite{gehring2016, kalchbrenner2016}. However, these works have found it is difficult to replace the target side of the model with CNN layers while maintaining high accuracy. The use of a recurrent target is especially important to track attentional coverage and ensure fluency.

Here, we propose a mixed model which uses an RNN layer at the bottom to both capture full-sentence context and perform attention, followed by a series of fully-connected (FC) layers applied on top at each timestep. The FC layers can be interpreted as a CNN without overlapping stride. Since each FC layer consists of a single matrix multiplication, it is $1/6^{\rm th}$ the cost of a GRU (or $1/8^{\rm th}$ an LSTM). Additionally, several of the speedups from Section~\ref{sec:speedups} can only be applied to the first layer, so there is strong incentive to only use a single target RNN.

To avoid vanishing gradients, we use ResNet-style skip connections \cite{he2016}. These allow very deep models to be trained from scratch and do not require any additional matrix multiplications, unlike highway networks \cite{srivastava2015}. With 5 intermediate FC layers, target timestep $i$ is computed as:
{\vskip -0.18in}
{\fontsize{10.0}{10.0}\selectfont
\begin{eqnarray*}
h^B_{i} &=& {\rm AttGRU}(h^B_{i-1}, x_i, S) \\
h^1_{i} &=& {\rm relu}(W^1h^B_i) \\
h^2_{i} &=& {\rm relu}(W^2h^1_i) \\
h^3_{i} &=& {\rm relu}(W^3h^2_i + h^1_i) \\
h^4_{i} &=& {\rm relu}(W^4h^3_i) \\
h^5_{i} &=& {\rm relu}(W^5h^4_i + h^3_i) \\
h^T_{i} &=& {\rm tanh}(W^Th^5_i)\ {\rm {\bf or}}\ {\rm GRU}(h^T_{i-1}, h^5_{i}) \\
y_i &=& {\rm softmax}(Vh^T_{i})
\end{eqnarray*}
{\vskip -0.09in}
}

We follow \citet{he2016} and only use skip connections on every other FC layer, but do not use batch normalization. The same pattern can be used for more FC layers, and the FC layers can be a different size than the bottom or top hidden layers. The top hidden layer can be an RNN or an FC layer. It is important to use {\tt relu} activations (opposed to {\tt tanh}) for ResNet-style skip connections. The GRUs still use {\tt tanh}.

\subsection{Model Results}
Results using the mixed RNN+FC architecture are shown in Table~\ref{table:model_results}, using all speedups. We have found that the benefit of using RNN+FC layers on the source is minimal, so we only perform ablation on the target. For the source, we use a 3-layer 512-dim bidi GRU in all models (S1)-(S6).

Model (S1) and (S2) are one and two layer baselines. Model (S4), which uses 7 intermediate FC layers, has similar decoding cost to (S2) while doubling the improvement over (S1) to 1.2 BLEU. We see minimal benefit from using a GRU on the top layer (S5) or using more FC layers (S6). In (E1) and (E2) we present 2 and 3 model ensembles of (S4), trained from scratch with different random seeds. We can see that the 2-model ensemble improves results by 0.9 BLEU, but the 3-model ensemble has little additional improvment. Although not presented here, we have found these improvement from decoder speedups and RNN+FC to be consistent across many language pairs.

All together, we were able to achieve a BLEU score of 38.3 while decoding at 100 words/sec on a single CPU core. As a point of comparison, \citet{googlenmt} achieves similar BLEU scores on this test set (37.9 to 38.9) and reports a CPU decoding speed of {\texttildelow}100 words/sec (0.2226 sents/sec), but parallelizes this decoding across 44 CPU cores. System (S7), which is our re-implementation of \citet{googlenmt}, decodes at 28 words/sec on one CPU core, using all of the speedups described in Section~\ref{sec:speedups}. \citet{zhou2016} has a similar computational cost to (S7), but we were not able to replicate those results in terms of accuracy.

Although we are comparing an ensemble to a single model, we can see ensemble (E1) is over 3x faster to decode than the single model (S7). Additionally, we have found that model (S4) is roughly 3x faster to train than (S7) using the same GPU resources, so (E1) is also 1.5x faster to train than a single model (S7).

\bibliography{nmt_decode}
\bibliographystyle{emnlp_natbib}

\end{document}